\documentclass[a4paper, 10pt, conference]{ieeeconf} 
\AtBeginDocument{\let\autocite\cite}

\IEEEoverridecommandlockouts                              

\usepackage{balance}
\usepackage{url}
\usepackage{cite}
\usepackage{graphicx} 
\usepackage{flushend}
\usepackage{booktabs}
\usepackage{tikz}

\begin{document}
\IEEEoverridecommandlockouts
\overrideIEEEmargins

\title{\LARGE \bf
  Diminishing Returns of Intelligence: The Non-Linear Relationship Between LLM Scale and User Perception in Short-Duration Open-Ended Social Human-Robot Interactions
}
\author{Amanda Rasille Røn Volf, and Morten Roed Frederiksen$^{1}$
  \thanks{{$^{1}$Amanda Rasille Røn Volf, and Morten Roed Frederiksen \tt\small mrof@itu.dk} are affiliated with the Computer Science Department of The IT-University of Copenhagen.}
}

\maketitle

\begin{abstract}
Large Language Models (LLMs) are increasingly used to drive embodied social agents, yet it remains unclear whether larger models improve user perception during brief human-robot encounters. This paper examines the effect of LLM parameter size on short-duration, open-ended social interactions with a robot interface. In a within-subjects study, 19 participants interacted with robot faces driven by Qwen3-VL models at 4B, 8B, and 30B parameters. Participants evaluated the interactions in terms of perceived intelligence, naturalness, enjoyment, and humor. Results showed no significant overall preference for the 30B model over the smaller variants, including no significant advantage over the 4B model in perceived naturalness or intelligence. A significant relationship between AI interaction frequency and intelligence rankings for the 30B model suggests that more experienced users may be more sensitive to differences in model capability. Overall, the findings indicate diminishing returns from model scaling in brief open-ended social HRI, where conversational flow, responsiveness, and socially appropriate behavior potentially matter as much as raw parameter count.
\end{abstract}
\section{Introduction}

The rapid evolution of Artificial Intelligence (AI) and robotic systems has accelerated the integration of autonomous agents into everyday human environments \cite{Shen2023ChatGPTAO, David2022TheAO}. Recent advances in Large Language Models (LLMs) have substantially expanded the cognitive and linguistic capabilities of AI systems \cite{Kasneci2023ChatGPTFG}. As a result, research increasingly explores how high-capacity LLMs can be embedded into physical robotic platforms to support richer social interaction \cite{Zhang2023LargeLM}. However, the psychological impact of deploying highly capable LLMs in embodied social robots remains under-explored, particularly with respect to whether increases in model scale translate into improved user perception during interaction.

Previous work in Human-Robot Interaction has shown that multimodal communication cues, including morphology, locomotion, and non-verbal behavior, can shape the affective impact of robots \cite{Shen2023ChatGPTAO, Frederiksen2020OnTC, Traeger2020VulnerableRP, Frederiksen2019ASC}. These design elements are especially important in high-touch domains such as elderly care, hospitality, and domestic assistance \cite{Bartneck2006TheIO, deKervenoael2020LeveragingHI}. In such contexts, successful interaction requires not only technical competence, but also social intelligence that supports user acceptance and trust \cite{Traeger2020VulnerableRP, Rossi2017HowTT}. LLMs may enable more intuitive natural-language communication \cite{Sonawani2024SiSCoSS, Kobzarev2025GestLLMAH, Murogaki2024EnhancingMD}; however, it remains unclear whether maximizing model capacity is always beneficial for human rapport.

Central to this research is the question of whether perceived intelligence can outweigh other interaction factors such as gesturing, morphology, timing, and social behavior. If intelligence is the primary driver of interaction quality, then social robot design may need to prioritize increasingly capable language models. If not, then smaller models may be sufficient for many short-duration encounters. By focusing on brief service-style interactions, we investigate whether the perceived benefits of LLM scaling are noticeable to users during short, non-repeated social exchanges.

\begin{figure}[t]
\centering
\includegraphics[width=0.485\textwidth]{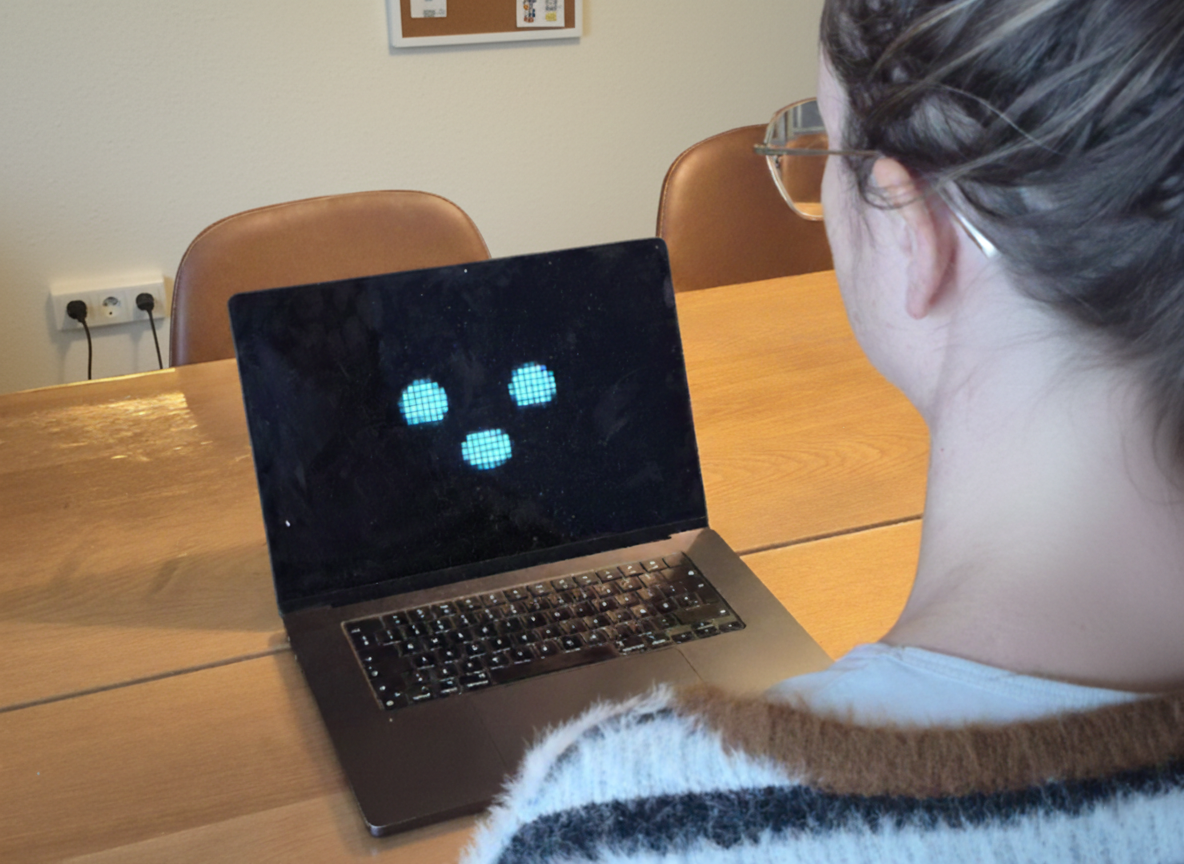}
\caption{
Experimental setup used for the open-ended interactions. Participants engaged with three visually distinct robot faces, each driven by a different LLM size. Facial colors were used to distinguish model conditions during the post-experiment questionnaires. The image was captured during initial setup testing and the background was removed for clarity.
}
\label{fig:experimental_setup}
\end{figure}

This paper investigates the extent to which LLM parameter scale influences social human-robot interaction, focusing on perceived intelligence, trust, likability, naturalness, enjoyment, and humor. We report an initial between-subjects pilot study followed by a within-subjects main study (N=19). In the main study, participants interacted with robot interfaces driven by three parameter sizes of the Qwen3-VL model: 4B, 8B, and 30B. Each interaction was open-ended and lasted two minutes. Participants completed pre- and post-interaction questionnaires, including adapted Godspeed and NARS-inspired measures, and ranked the three robot configurations after completing all interactions.

Our findings suggest diminishing perceptual returns from model scaling in short-duration open-ended interactions. Although the pilot study indicated that smaller models could be perceived as slightly more humorous, the main study showed no significant overall preference for the 30B model over the smaller variants in perceived intelligence, naturalness, enjoyment, or humor. The 30B model's enjoyment ranking was significantly associated with humor ranking (p=.017), suggesting that successful social behavior may be more important than raw parameter count in brief interactions. At the same time, the 4B model remained competitive in perceived intelligence for the average participant.

These results suggest that computational scaling does not automatically produce a stronger social connection, and user perception appears to depend strongly on conversational flow, responsiveness, and social behavior. Consequently, this work argues that future research should complement model scaling with the optimization of social interaction behaviors, including humor, empathy, and timing.
\section{Pilot Study}

An initial pilot study with 15 participants compared an instruction-tuned Alpaca 7B model and a Llama 13B base model in a between-subjects text-based interaction. The pilot suggested that smaller models could be perceived as equally or more intelligent when they produced faster or more socially engaging responses. Because of its limited sample size and different model families, the pilot was used only to motivate the main within-subjects study reported below.

\section{Method}

We conducted a within-subjects study with 19 participants aged 18--58. Each participant interacted with three robot configurations driven by Qwen3-VL models with 4B, 8B, and 30B parameters. Each interaction lasted two minutes. The order of model presentation was randomized across participants.

The robot interface consisted of an animated face displayed on a laptop. The face blinked, smiled, tracked the participant using a camera, and responded through synthesized speech. Speech input was captured through a microphone and transcribed using Whisper. Each model condition was associated with a distinct face color to help participants identify the three interactions during the ranking task. The color-to-model mapping was kept constant, which improved recall but introduced a potential visual confound.

All models used the same system prompt:

\begin{quote}
You are a friendly, funny and social robot that loves having conversations! Keep your responses short and friendly. Use a conversational tone and be engaging. Default to replies under 2 sentences unless the user asks for detail.
\end{quote}

After each interaction, participants completed questionnaire items adapted from Godspeed-style measures, using 7-point Likert scales for qualities such as intelligence, sense-making, understanding, humor, smiling, responsiveness, and response speed. After all three interactions, participants ranked the robots by enjoyment, naturalness, intelligence, and humor. A pre-experiment questionnaire collected participant-level covariates, including self-reported frequency of AI interaction, caution toward new technology, and expectations about whether technology usually functions correctly.

To quantify backend timing differences, we ran an additional generation-speed test. Each model was queried ten times with the same test script and prompt. We recorded words per second, tokens per second, time to first token (TTFT), generation time, and total request time. These measurements capture backend model performance only, not full end-to-end conversational latency.

\section{Results}

\begin{figure*}[h]
\centering
\includegraphics[width=1\textwidth]{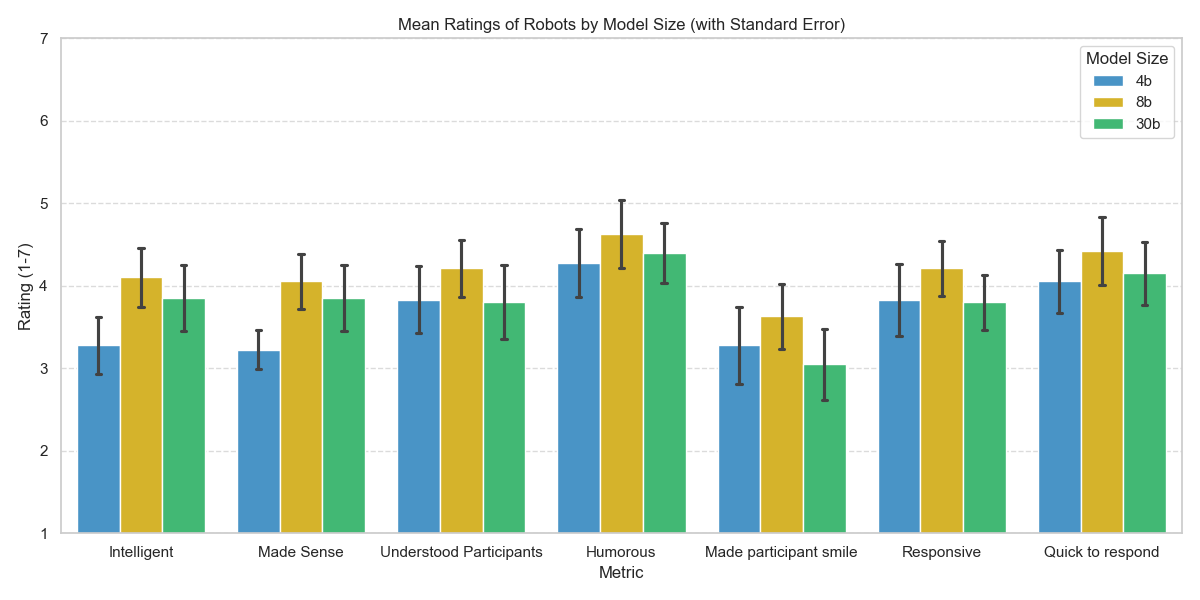}
\caption{Comparison of the participants' quality ratings between LLM scales. The high degree of overlap across perceived intelligence, coherence, understanding, humor, smiling, responsiveness, and response speed indicates that model parameter count did not substantially alter user perception during two-minute open-ended dialogues.}
\label{fig:ratings}
\end{figure*}

Across the post-interaction ratings, participants showed no clear overall preference for the largest model. As shown in Fig.~\ref{fig:ratings}, ratings overlapped substantially across model sizes for perceived intelligence, sense-making, understanding, humor, smiling, responsiveness, and response speed.

The backend timing test showed that responsiveness did not scale monotonically with parameter count. The 4B model had the shortest total request time (2.03 s), followed by the 30B model (2.32 s), while the 8B model was slowest (2.71 s). TTFT was nearly identical across models, ranging from 0.23 s to 0.24 s, suggesting that timing differences mainly appeared during generation rather than initial response onset.

Two exploratory correlations provide additional context. Participants' self-reported frequency of AI interaction was negatively correlated with intelligence ranking for the 30B model (r(18)=-.621, p=.005), suggesting that more experienced users may have been more sensitive to model capability. For the 30B model, humor ranking was positively correlated with enjoyment ranking (r(18)=.539, p=.017), indicating that successful humor may be especially important for the perceived quality of larger-model interactions.

\begin{table}[t]
\centering
\caption{Backend timing by model. Values are mean $\pm$ SD over ten requests.}
\label{tab:latency}
\begin{tabular}{lccc}
\hline
Model & TTFT (s) & Gen. (s) & Total (s) \\ \hline
4B  & $0.23 \pm 0.02$ & $1.80 \pm 0.01$ & $2.03 \pm 0.02$ \\
8B  & $0.24 \pm 0.03$ & $2.46 \pm 0.01$ & $2.71 \pm 0.03$ \\
30B & $0.24 \pm 0.03$ & $2.08 \pm 0.02$ & $2.32 \pm 0.02$ \\ \hline
\end{tabular}
\end{table}

\subsection{The Humor of Small Models}

The findings from the pilot study suggest a counter-intuitive trend: the smaller 7B model was perceived as more intelligent, a result that may have been influenced by its perceived humor, instruction-tuned behavior, and the subjective experience of  responsiveness. This suggests that the perceived intelligence of a language model in a social robotics context is not purely a function of its reasoning capabilities or parameter count, but may also depend on whether the overall interaction feels timely, engaging, and socially coherent. However, the backend timing test conducted for the main study showed that the 30B model was slower than the 4B model but faster
than the 8B model. This indicates that model size, backend timing, and
perceived responsiveness should potentially not be treated as a linear trade-off.

The perceived humor in smaller models may ironically be a byproduct of their technical limitations. Humor often thrives where a response is unexpected or slightly out of place within the social context. Because smaller models possess a less robust contextual understanding compared to their larger counterparts, they are statistically more likely to produce happy accidents. In the eyes of the user, these unexpected pivots are often interpreted as intentional wit or a sign of a distinct personality. Since humor is traditionally viewed as a high-level cognitive trait, these events in contextual mapping could mask the model's smaller scale, allowing it to rate equal to or higher than denser models in perceived intelligence.

\subsection{Similar Outcomes for Short-Duration Interactions}

While the results of the main study were less divergent than the pilot, the consistency in ratings across the model spectrum indicates that smaller models are remarkably viable for social robotics. It appears feasible for robots powered by smaller LLMs to appear both intelligent and engaging, provided the interaction remains within a specific temporal window.

Our data indicates these trends for shorter duration conversations, which aligns perfectly with the practical requirements of the service robot industry. In applications such as robot taxis, waiter robots, or delivery units, interactions are naturally brief and goal-oriented. In these contexts, there is a significant economic and computational advantage to utilizing smaller models. By reducing computational requirements and deployment overhead, developers may be able to provide a responsive, intelligent-feeling interface without relying on the largest available models. However, our backend timing test also shows that responsiveness is deployment-specific and does not scale strictly
with parameter count.

It is also important to note that the interactions in this study were open-ended and lacked a specific goal. In a scenario where the robot has a strict task to achieve (e.g., resolving a complex customer service complaint), the superior reasoning and instruction-following capabilities of larger models would likely become a necessity. For small-talk and simple guidance, however, the smaller-model range may be sufficient. Therefore, the findings should not be interpreted as evidence that smaller models are generally equivalent to larger models for all social robotics applications. Rather, they suggest that in short, open-ended, low-stakes encounters, participants may not strongly benefit from the additional reasoning capacity of larger models. Longer interactions, memory-dependent dialogue, and goal-directed service tasks may reveal advantages of larger models that were not observable in the present two-minute interaction format.

\subsection{Future Potential}

A notable constraint of this study was that the models did not receive specific behavioral instructions, operating primarily on standard system prompts. This opens a significant avenue for future research: investigating if the perceived intelligence of small models be artificially boosted by explicitly instructing them to be humorous. If the humor observed in our pilot could be systematized through prompt engineering, the gap between small-scale efficiency and large-scale perception might close even further.
An indicated limitation concerns the use of different face colors to distinguish model conditions. Although the colors were intended only as memory aids for the ranking questionnaire, visual appearance can influence social perception in Human-Robot Interaction. Future studies should either use the same face color across all model conditions or counterbalance the color-to-model mapping across participants to eliminate color as a potential confound. A further limitation is that the timing test measured backend model performance rather than full end-to-end conversational latency. Participants experienced the complete interaction pipeline, including speech-to-text transcription, model inference, text-to-speech synthesis, and the timing of the robot's spoken response. Future studies should instrument the full pipeline to determine whether perceived responsiveness is better explained by end-to-end turn latency than by backend generation speed alone.
Furthermore, the rise of agentic systems suggests a hybrid path forward.In these architectures, smaller or more computationally efficient models could handle brief, context-aware processing and routine dialogue, while a larger thinking model is reserved for complex problem-solving. Future research should explore dynamic model switching, where a system identifies the conversational context in real-time and toggles between model sizes to optimize for both speed and depth. This approach would leverage the inherent social advantages of smaller models while maintaining the intellectual safety net of larger ones.
\section{Conclusion}
This study examined whether increasing the parameter size of an LLM improves user perception in short-duration, open-ended social Human-Robot Interactions. Across a within-subjects study with 19 participants comparing Qwen3 4B, 8B, and 30B models, we found no statistically significant overall preference for the largest model in perceived intelligence, naturalness, enjoyment, or humor. These findings suggest that, within brief service-style encounters, increasing model scale alone does not necessarily translate into a noticeably better user experience and that factors such as responsiveness, conversational flow, and socially salient behaviors may be as important as, or more important than, raw model capacity. The correlation between AI familiarity and intelligence rankings for the 30B model suggests that more experienced users may be better able to detect subtle differences in model capability. However, for the average participant in a short interaction, smaller models remained competitive with larger models.

The results have practical implications for the design of social robots in domains such as hospitality, transportation, and customer assistance, where interactions are often brief and temporally constrained. In such contexts, optimized smaller models may provide a favorable trade-off between perceived intelligence, deployment cost, and computational requirements, while responsiveness should be evaluated empirically for each deployment rather than assumed from parameter count alone. Future work should investigate longer interactions, task-oriented scenarios, prompt-engineered social behaviors, and dynamic architectures that combine fast small models with larger models for complex reasoning when needed.

\balance

\bibliography{bibliography}

@article{Shen2023ChatGPTAO,
  title={ChatGPT and Other Large Language Models Are Double-edged Swords.},
  author={Yiqiu Shen and Laura Heacock and Jonathan Elias and Keith D Hentel and Beatriu Reig and George Shih and Linda Moy},
  journal={Radiology},
  year={2023},
  pages={
          230163
        }
}

@article{Frederiksen2020OnTC,
  title={On the causality between affective impact and coordinated human-robot reactions},
  author={Morten Roed Frederiksen and Kasper St{\o}y},
  journal={2020 29th IEEE International Conference on Robot and Human Interactive Communication (RO-MAN)},
  year={2020},
  pages={488-494}
}

@article{deKervenoael2020LeveragingHI,
  title={Leveraging human-robot interaction in hospitality services: Incorporating the role of perceived value, empathy, and information sharing into visitors’ intentions to use social robots},
  author={Ronan de Kervenoael and Rajibul Hasan and Alexandre Schwob and Edwin Goh},
  journal={Tourism Management},
  year={2020},
  volume={78},
  pages={104042}
}

@article{David2022TheAO,
  title={The acceptability of social robots: A scoping review of the recent literature},
  author={Dayle David and Pierre Th{\'e}rouanne and Isabelle Milhabet},
  journal={Comput. Hum. Behav.},
  year={2022},
  volume={137},
  pages={107419}
}

@inproceedings{Rossi2017HowTT,
  title={How the Timing and Magnitude of Robot Errors Influence Peoples' Trust of Robots in an Emergency Scenario},
  author={Alessandra Rossi and Kerstin Dautenhahn and Kheng Lee Koay and Michael L. Walters},
  booktitle={International Conference on Software Reuse},
  year={2017}
}

@article{Murogaki2024EnhancingMD,
  title={Enhancing Multi-Person Dialogue with Large Language Models: A Structured Approach to Natural Communication},
  author={Takumi Murogaki and Toshikazu Nishimura},
  journal={Proceedings of the 2024 8th International Conference on Natural Language Processing and Information Retrieval},
  year={2024}
}

@article{Kobzarev2025GestLLMAH,
  title={GestLLM: Advanced Hand Gesture Interpretation via Large Language Models for Human-Robot Interaction},
  author={Oleg Kobzarev and Artem Lykov and Dzmitry Tsetserukou},
  journal={2025 20th ACM/IEEE International Conference on Human-Robot Interaction (HRI)},
  year={2025},
  pages={1413-1417}
}

@article{Sonawani2024SiSCoSS,
  title={SiSCo: Signal Synthesis for Effective Human-Robot Communication Via Large Language Models},
  author={Shubham D. Sonawani and Fabian Clemens Weigend and Heni Ben Amor},
  journal={2024 IEEE/RSJ International Conference on Intelligent Robots and Systems (IROS)},
  year={2024},
  pages={7107-7114}
}

@article{Bartneck2006TheIO,
  title={The influence of people’s culture and prior experiences with Aibo on their attitude towards robots},
  author={Christoph Bartneck and Tomohiro Suzuki and Takayuki Kanda and Tatsuya Nomura},
  journal={AI \& SOCIETY},
  year={2006},
  volume={21},
  pages={217-230}
}

@article{Traeger2020VulnerableRP,
  title={Vulnerable robots positively shape human conversational dynamics in a human–robot team},
  author={Margaret Traeger and Sarah Strohkorb Sebo and Malte F. Jung and Brian Scassellati and Nicholas A. Christakis},
  journal={Proceedings of the National Academy of Sciences of the United States of America},
  year={2020},
  volume={117},
  pages={6370 - 6375}
}

@article{Kasneci2023ChatGPTFG,
  title={ChatGPT for good? On opportunities and challenges of large language models for education},
  author={Enkelejda Kasneci and Kathrin Se{\ss}ler and Stefan K{\"u}chemann and Maria Bannert and Daryna Dementieva and Frank Fischer and Urs Gasser and George Louis Groh and Stephan G{\"u}nnemann and Eyke H{\"u}llermeier and Stephan Krusche and Gitta Kutyniok and Tilman Michaeli and Claudia Nerdel and J{\"u}rgen Pfeffer and Oleksandra Poquet and Michael Sailer and Albrecht Schmidt and Tina Seidel and Matthias Stadler and Jochen Weller and Jochen Kuhn and Gjergji Kasneci},
  journal={Learning and Individual Differences},
  year={2023}
}

@article{Frederiksen2019ASC,
  title={A Systematic Comparison of Affective Robot Expression Modalities},
  author={Morten Roed Frederiksen and Kasper St{\o}y},
  journal={2019 IEEE/RSJ International Conference on Intelligent Robots and Systems (IROS)},
  year={2019},
  pages={1385-1392}
}

@article{Zhang2023LargeLM,
  title={Large language models for human-robot interaction: A review},
  author={Ceng Zhang and Junxin Chen and Jiatong Li and Yanhong Peng and Ze-bing Mao},
  journal={Biomimetic Intelligence and Robotics},
  year={2023}
}
\bibliographystyle{IEEEtran}

\end{document}